\pdfoutput=1 
\documentclass[10pt, twoside]{procdoc2}

\usepackage{IEEEtrantools}
\usepackage{graphicx}
\usepackage{booktabs}
\usepackage{xcolor}
\usepackage{amsmath,amsfonts}
\usepackage{pifont}

\usepackage{tikz}

\definecolor{color1}{HTML}{DC653D}
\definecolor{color2}{HTML}{222A35}
\definecolor{color3}{HTML}{445469}
\definecolor{color4}{HTML}{8497B0}
\definecolor{color5}{HTML}{772E15}
\definecolor{color6}{HTML}{52883B}

\newcommand{\myline}[1]{\raisebox{2pt}{\tikz{\draw[-, #1, solid, line width = 1.3pt](0,0) -- (5mm,0);}}}

\newcommand{\mydashline}[1]{\raisebox{2pt}{\tikz{\draw[-, #1, dashed, line width = 1.3pt](0,0) -- (5mm,0);}}}

\newcommand{\mydashdotline}[1]{%
  \raisebox{2pt}{\tikz{\draw[-, #1, dash dot, line width=1.3pt](0,0)--(5mm,0);}}%
}

\usepackage[title]{appendix}

\usepackage{lipsum}
\usepackage{lineno}

\title{Probabilistic Multi-Layer Perceptrons for Wind Farm Condition Monitoring}
\author[1]{Filippo Fiocchi}
\author[2, *]{Domniki Ladopoulou}
\author[2, 3]{Petros Dellaportas}
\affil[1]{Department of Computer Science, University College London, London, United Kingdom}
\affil[2]{Department of Statistical Science, University College London, London, United Kingdom}
\affil[3]{Department of Statistics, Athens University of Economics and Business, Athens, Greece}
\affil[*]{Corresponding author;  e-mail: domna.ladopoulou.22@ucl.ac.uk}
\date{}

\makeatletter
\def\runauthor{F. Fiocchi, D. Ladopoulou and P. Dellaportas}
\def\runtitle{Probabilistic Multi-Layer Perceptrons for Wind Farm Condition Monitoring}
\makeatother

\begin{document} 

\thispagestyle{empty}

\twocolumn[
  \maketitle
  \begin{onecolabstract}
We provide a condition monitoring system for wind farms,
based on normal behaviour modelling using a probabilistic multi-layer perceptron with transfer learning via fine-tuning.
The model predicts the output power of the wind turbine under normal behaviour  based on features retrieved from supervisory control and data acquisition (SCADA) systems.  Its advantages are that  (i) it can be trained with SCADA data of at least a few years, (ii) it can incorporate all SCADA data of all wind turbines in a wind farm as features, (iii) it assumes that the output power follows a normal density with heteroscedastic variance and (iv) it can predict the output of one wind turbine  by borrowing strength from the data of all other wind turbines in a farm. Probabilistic guidelines for condition monitoring are given via a cumulative sum (CUSUM) control chart, which is specifically designed based on a real-data classification exercise and, hence, is adapted to the needs of a wind farm. We illustrate  the performance of our model in a real SCADA data example which provides evidence that it outperforms other probabilistic  prediction models.
  \end{onecolabstract}
]

\begin{table}[ht]
\centering
\begin{tabular}{l|l}
\textit{Abbreviation} & \textit{Definition} \\\midrule\midrule
CP              & Coverage Probability \\
CUSUM           & Cumulative Sum \\
LPMLP           & Large Probabilistic Multi-Layer Perceptron \\
LSTM            & Long Short-Term Memory \\
MAE             & Mean Absolute Error \\
MCE             & Maximum Calibration Error \\
NMAE            & Normalized Mean Absolute Error \\
NRMSE           & Normalized Root Mean Square Error \\
PMLP            & Probabilistic Multi-Layer Perceptron \\
RBF             & Radial Basis Function \\
ReLU            & Rectified Linear Unit \\
RMSE            & Root Mean Square Error \\
SCADA           & Supervisory Control and Data Acquisition \\
\end{tabular}
\label{tab:acronyms}
\end{table}

\section{Introduction} \label{intro}
Due to the environmental and sustainability benefits, many energy producers have shifted their attention towards wind energy; one of the cleanest and fastest-growing sources of renewable energy. However, the severe weather conditions and the often remote locations of wind farms lead to both high operational failure rates and high maintenance costs that account for $20-30\%$ of the total cost related to power generation  \citep{reference2}. There is therefore a  need to reduce the high operational and maintenance costs of wind turbines via a cost-effective and high-precision  condition monitoring system.  Such a system is 
crucial for an early fault detection and thus minimisation of the prolonged wind turbine downtime. 

While specific monitoring sensors can be employed to detect incoming faults, wind farm operators are sceptical of employing expensive technology without direct economic justification \citep{reference1}.  Most wind farms have already been equipped with supervisory control and data acquisition (SCADA) systems. These systems record the operational status of individual wind turbines and all their components, such as wind speed, power, rotor speed and blade pitch angles. They produce large amounts of data, which are collected at a high frequency, for example, one Hz, and recorded as a $10$-minute averaged interval. Since these systems are already installed, monitoring performance and predicting early faults by extracting features from SCADA data is a widely researched cost-effective measure of failure prevention. 

A typical approach to the problem is to produce a model that predicts the output power of a wind turbine under normal behaviour based on features retrieved from SCADA data; see, for example, \cite{reference3}. Then, one could monitor the wind turbine operation by using the model to inspect whether the observed and model-predicted output powers deviate in some statistical sense. Although this is a sensible, well-known approach, practical and effective condition monitoring of a wind farm requires a modelling perspective with four particular characteristics, outlined below.

First, to accommodate weather fluctuations, the model needs to be able to scale well with the data size since it requires to be trained for at least one calendar year which is $52,560$ $10$-minute intervals.  Second, it should be able to extract all the wealth of information available in the SCADA system  which is expressed in a few dozen features per wind turbine per time interval.
Third, the model needs to provide probabilistic predictions in the form of a predictive density and not only point estimates of output power so that a proper probabilistic assessment of the deviation of the observed from the expected output power can be performed.  The heteroscedastic nature of the predictions is also necessary to properly model the wind turbine output power because low and high wind speeds produce low variance power distributions.
This immediately requires that the predictive density has sufficiently good coverage probabilities in out-of-sample data scenarios.  Last but not least, the model should be able to deal with the peculiarities of wind farm SCADA data recording  in the following sense.  It is common that historical SCADA data may have many missing data in a particular wind turbine because it was out-of-order for a long period of time or it has been recently installed.  A good model should be able to predict the output power of this wind turbine with an inferential procedure that borrows strength from the features and output power of all other wind turbines in the wind farm.

We develop a condition monitoring system based on a model that has all the four necessary characteristics described above based on a probabilistic multi-layer perceptron (PMLP) with transfer learning via fine-tuning.  
We assume that the output power follows a normal distribution with input-dependent mean and input-dependent variance which form our predictive density for each $10$-minute interval. 
For the probabilistic condition monitoring we propose the use of cumulative sum (CUSUM) control charts.
{We illustrate its performance in a real-data application by comparing it with two other probabilistic models: a sparse Gaussian process and a Bayesian neural network, as shown in Table \ref{table:results}.  We found that  our model predicts better with respect to root mean square error (RMSE), mean absolute error (MAE) and maximum calibration error (MCE).}  We demonstrate that by transfer learning we can use information from all turbines in a wind farm to improve the prediction of the output power of a single wind turbine. 
We also provide a real data example in which our proposed condition monitoring system expressed via a CUSUM control chart reveals an early warning in a particular wind turbine failure. 

The rest of this paper is organized as follows.  Section \ref{related-work} contains related work, Section \ref{methodology} presents the proposed methodology, Section \ref{experiments} presents the empirical application, Section \ref{discussion}  discusses further developments and Section \ref{conclusion} concludes with our key findings.

\section{Related Work} \label{related-work}

Our proposed modelling perspective is based on the notion of normal behaviour modelling, which attempts to produce, conditional on SCADA input features,  predictions of the output power of a wind turbine under healthy conditions and then diagnose possible anomalous performance by comparing the predicted and the observed output powers.  This is an unsupervised learning strategy which differs from another strand of the literature that uses supervised learning and exploits fault instances obtained from operational and events files from a SCADA system.  Thus, in this Section, we will only present related work from articles on modelling normal behaviour.  The more relevant to our work papers that use probabilistic predictions will be presented in subsection \ref{sec: probabilistic}.

Normal behaviour modelling studies vary in their selection of the input features used to identify faults.  Predictions are performed to the output power \citep{reference3, reference6, reference7, kusiak-2011, schlechtingen-2013}, the generator's temperature \citep{reference1, reference14, reference8}, the gearbox bearings' temperature \citep{reference9, reference10, reference1, reference8, reference15}, or multiple output variables \citep{reference8, reference11, reference19}.  
\subsection{Non-probabilistic normal behaviour modelling}

A series of articles \citep{reference1,reference8,reference9, reference15,reference19,reference10}, proposed deep neural networks to model operational characteristics such as the temperature of the gearbox bearings, cooling oil and winding temperature of the wind turbine. 
Autoencoders have been proposed by \cite{reference12} and \cite{yang-2021} whereas generative adversarial networks combined with autoencoders have been used by \cite{reference17}.  
Other methods include \citep{reference14}, who predicted the generator’s temperature with a nonlinear state estimate technique, and \citep{schlechtingen-2013}, who proposed an adaptive neuro-fuzzy inference system to predict the output power.

\subsection{Probabilistic normal behaviour modelling}
\label{sec: probabilistic}

Due to their intrinsic probabilistic nature,  Gaussian processes are  common candidates for probabilistic normal behaviour modelling, see for example \cite{reference6, reference3, reference7}.  The large impediment to their widespread application, namely their computational complexity which increases cubically with the data sample size, is usually being dealt with variational inference with inducing points \citep{titsias2009variational}.  However, different input features increase the hyper-parameter dimension since proper implementation requires different kernels in each dimension, which creates both convergence and kernel identification problems.  
Finally, although the issue of heteroscedasticity can be solved by assuming that the error process also follows a Gaussian process \citep{lazaro2011variational}, it is extremely difficult to resolve all three issues above simultaneously.  Finally, although multi-task Gaussian processes is a modelling perspective that simultaneously exploits the data of all wind turbines, the computational complexity increases quickly and treating missing values from one or more wind turbines is not possible. 
The works of \cite{reference18, reference20} have incorporated noise heteroscedasticity whereas the trade-off between heteroscedasticity and computational complexity has been considered by \cite{reference3}.
Recently, \cite{deng-2024} addressed the necessity of a probabilistic setting for power curve estimation by incorporating prior distributions on the weight parameters of a  long short-term memory (LSTM) Bayesian neural network. They improve their model by using a temporal convolutional neural network for temporal feature learning and an embedding layer to map discrete features (e.g., integer years) to dense vectors. The predictions are obtained via Monte Carlo sampling.
Finally, \cite{zhang-2020} considered the problem of predicting output power conditional on future wind speed and direction forecasts and adopted a modelling approach that closely resembles ours.  They employed an improved deep mixture density network by transforming the output energy to the interval $(0,1)$ and obtained probabilistic power predictions by assuming that it follows a mixture of Beta distributions.  As is similarly emphasized in our study, they highlighted the fact that informed decision-making is one of the key benefits of a probabilistic approach to wind power prediction. 

\section{Proposed Methodology} \label{methodology}

\begin{figure}[ht]
    \centering
    \includegraphics[width=1\columnwidth]{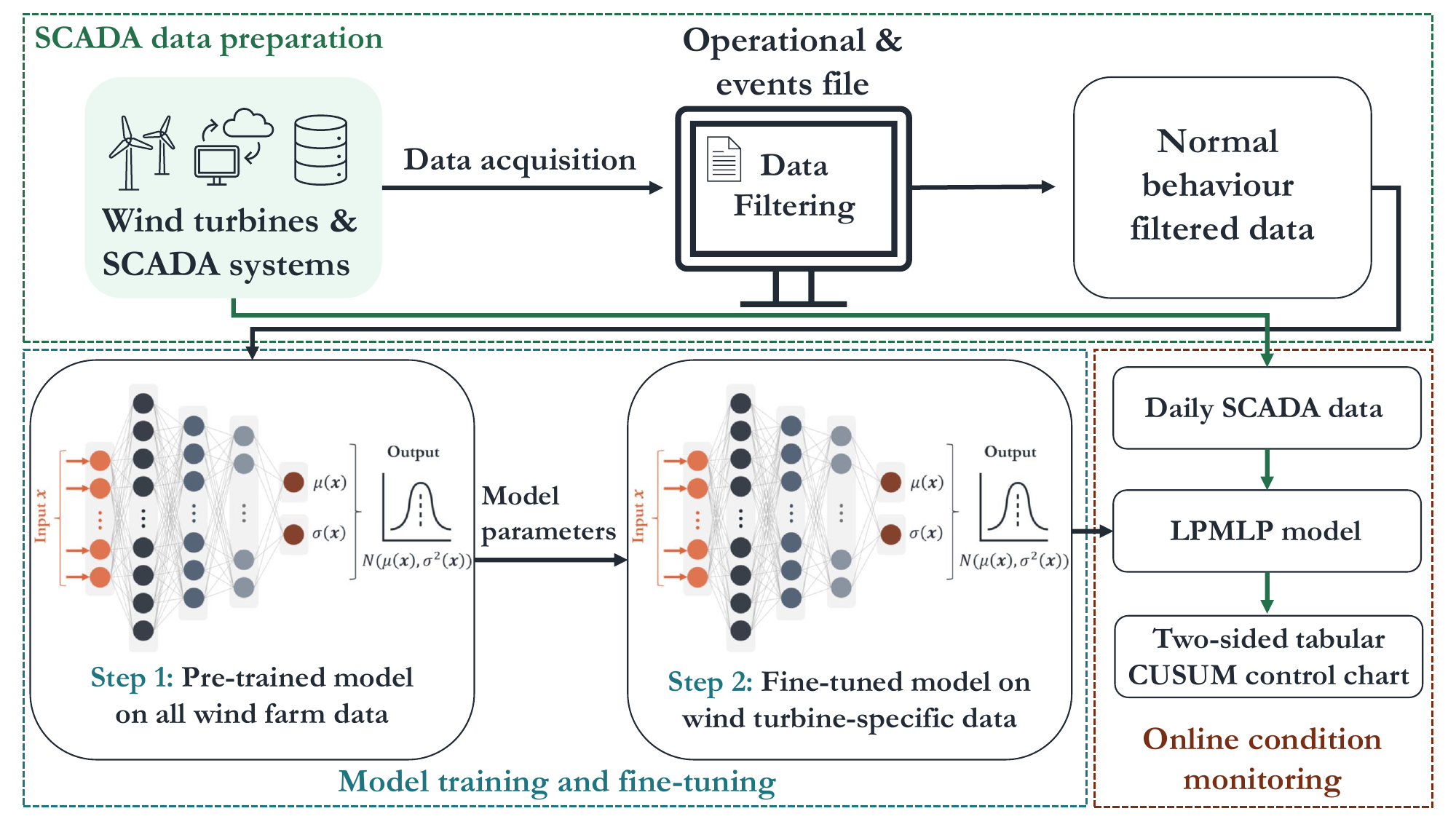}
    \caption{Overall flowchart of the proposed probabilistic condition monitoring system. LPMLP: Large probabilistic multi-layer perceptron presented in Section \ref{prob-mlp-ft}.}
    \label{fig:proposed_method}
 \end{figure}

We present a modelling perspective guided by the need to provide a condition monitoring system that has immediate practical applications in any wind farm.  We first describe a fully connected deep neural network that predicts both the mean and the variance of output power at a particular $10$-minute interval.  The computational power of the neural network, together with the stochastic heteroscedastic output power, allows the incorporation of many features obtained from large SCADA datasets that achieve good training based on as many as possible environmental and operational conditions and a statistically sound monitoring system based on both the mean and the variance of the power output.  The final ingredient in our model that is hugely important and necessary in realistic condition monitoring systems is the incorporation of transfer learning which is simultaneously trained in all wind turbines in a wind farm.  Finally, we present a realisation of our proposed probabilistic condition monitoring system based on a CUSUM control chart.  The overall flowchart of the proposed probabilistic condition monitoring
system is presented in Fig. \ref{fig:proposed_method}.

\subsection{Probabilistic multi-layer perceptron} \label{prob-mlp}

To model a heteroscedastic output noise we assume that for each  $10$-minute interval the power output of a wind turbine $y$ conditioned on the SCADA input features $\mathbf{x} \in \Re^{d_0}$ follows a normal distribution with unknown mean and variance that both depend on $\mathbf{x}$, so  
$y \sim \mathcal{N}(\mu(\mathbf{x}), \sigma^2(\mathbf{x}))$.
We will approximate $\mu(\mathbf{x})$ and $\sigma(\mathbf{x})$  by a PMLP with branching prediction heads. We now describe the architecture of our proposed model.

Given an input vector $\mathbf{x}= (x_{1},x_{2},\ldots,x_{d_0})$ that represents the $d_0$ SCADA input features at a particular time interval, the predicted PMLP power output $\hat{y} \in \Re$ follows a normal density with parameters $\mu(\mathbf{x})$ and $\sigma^2(\mathbf{x})$ which are continuous functions of the inputs  $\mathbf{x}$. 
In fact, our proposed PMLP models $\mu(\mathbf{x})$ as a linear piecewise function of $\mathbf{x}$  and $\sigma(\mathbf{x})$  as a nonlinear piecewise function of $\mathbf{x}$.

For a number of $k$ hidden layers  with widths $d_1,d_2,\ldots,d_{k}$ and input and output dimensions $d_0$ and $d_{k+1}=1$ respectively,
we adopt the linear functions $g_i : \Re^{d_{i-1}} \rightarrow \Re^{d_i}$ for $i=1,\ldots,k$ defined as 
 $$
 g_i(\mathbf{x}) = \mathbf{W}_i \mathbf{x} + \mathbf{b}_i,
 $$
 where $\mathbf{W}_i \in \Re^{d_{i} \times d_{i-1}}$, $\mathbf{x} \in \Re^{d_{i-1}}$ and $\mathbf{b}_i \in \Re^{d_{i}}$.  
 Furthermore, we adopt the Rectified Linear Unit (ReLU) function $ReLU_i : \Re^{d_i} \rightarrow \Re^{d_i}$ defined as
 {\small$$
 ReLU_i(\mathbf{x}) = (\max\{0,x_1\},\max\{0,x_2\},\ldots,\max\{0,x_{d_i}\}),
 $$}
 and the Softplus function  $S : \Re \rightarrow \Re$ defined as 
 \begin{equation}
 S(x) = \log (1+e^{x}) + \delta \label{softplus}
 \end{equation} 
 for a small $\delta >0$ used for numerical stability needed when $x$ becomes very small.
 These three functions are the basic ingredients of our PMLP, but we note that the functions $ReLU$ and $S$ can be replaced by any other appropriate alternative. 
 
 We now define the branching structure of the PMLP. 
 We use two sets of functions $g$ and $ReLU$, corresponding to predictions of 
 $\mu(\mathbf{x})$ and $\sigma(\mathbf{x})$
 and indexed by the superscripts $\mu$ and $\sigma$ respectively.   Different functions $g$ have different sets of parameters  $\mathbf{W}_i, \mathbf{b}_i$ where by different functions $ReLU$ is meant that they operate on different dimensions $d_i$.  We further assume that there exist two corresponding numbers of hidden layers $k_\mu$ and $k_\sigma$ with widths $d^\mu_1,d^\mu_2,\ldots,d^\mu_{k_\mu}$ and  $d^\sigma_1,d^\sigma_2,\ldots,d^\sigma_{k_\sigma}$.  
 Then, the output ${\hat y}$ of the PMLP  with $k_\mu$ and $k_\sigma$ hidden layers is represented as
 {\small\begin{align} 
 \mu(\mathbf{x}) & = g^\mu_{k_\mu+1} \circ ReLU^\mu_{k_\mu} \circ g_{k_\mu}^\mu \circ  \cdots  g^\mu_2 \circ ReLU^\mu_1 \circ g^\mu_1(\mathbf{x}) \nonumber \\
 \sigma(\mathbf{x}) & = S \circ g^\sigma_{k_\sigma+1} \circ ReLU^\sigma_{k_\sigma} 
 \circ g^\sigma_{k_\sigma}  \circ
 \cdots   g^\sigma_2 \circ ReLU_1^\sigma \circ g^\sigma_1(\mathbf{x}) \nonumber \\
 {\hat y} & \sim \mathcal{N}(\mu(\mathbf{x}), \sigma^2(\mathbf{x})) \label{model}
\end{align}}
 where $\circ$ denotes the function decomposition operator.

The branching mode of the PMLP is achieved by setting a layer $k^*$ such that $1 \leq k^* \leq \min \{k_\mu,k_\sigma\}$ and setting 
 \begin{align*}
ReLU^\mu_i(\mathbf{x}) & = ReLU^\sigma_i(\mathbf{x}) \\
g^\mu_i(\mathbf{x}) & = g^\sigma_i(\mathbf{x}) \\
d^\mu_i &=  d^\sigma_i
\end{align*}
for all $1 \leq i \leq k^*$.  Thus, before the \( k^* \)-th layer there is a common deep neural network; after the \( k^* \)-th layer, the network branches into two paths for predicting \( \mu(\mathbf{x}) \) and \( \sigma(\mathbf{x}) \) with each path having its own sequence of hidden layers with potentially different widths and depths.
Note that the total sum of hidden layers is $k_{total}= k_\mu+k_\sigma - k^*$.

\begin{figure}[htbp]
    \centering
    \includegraphics[width=1\columnwidth]{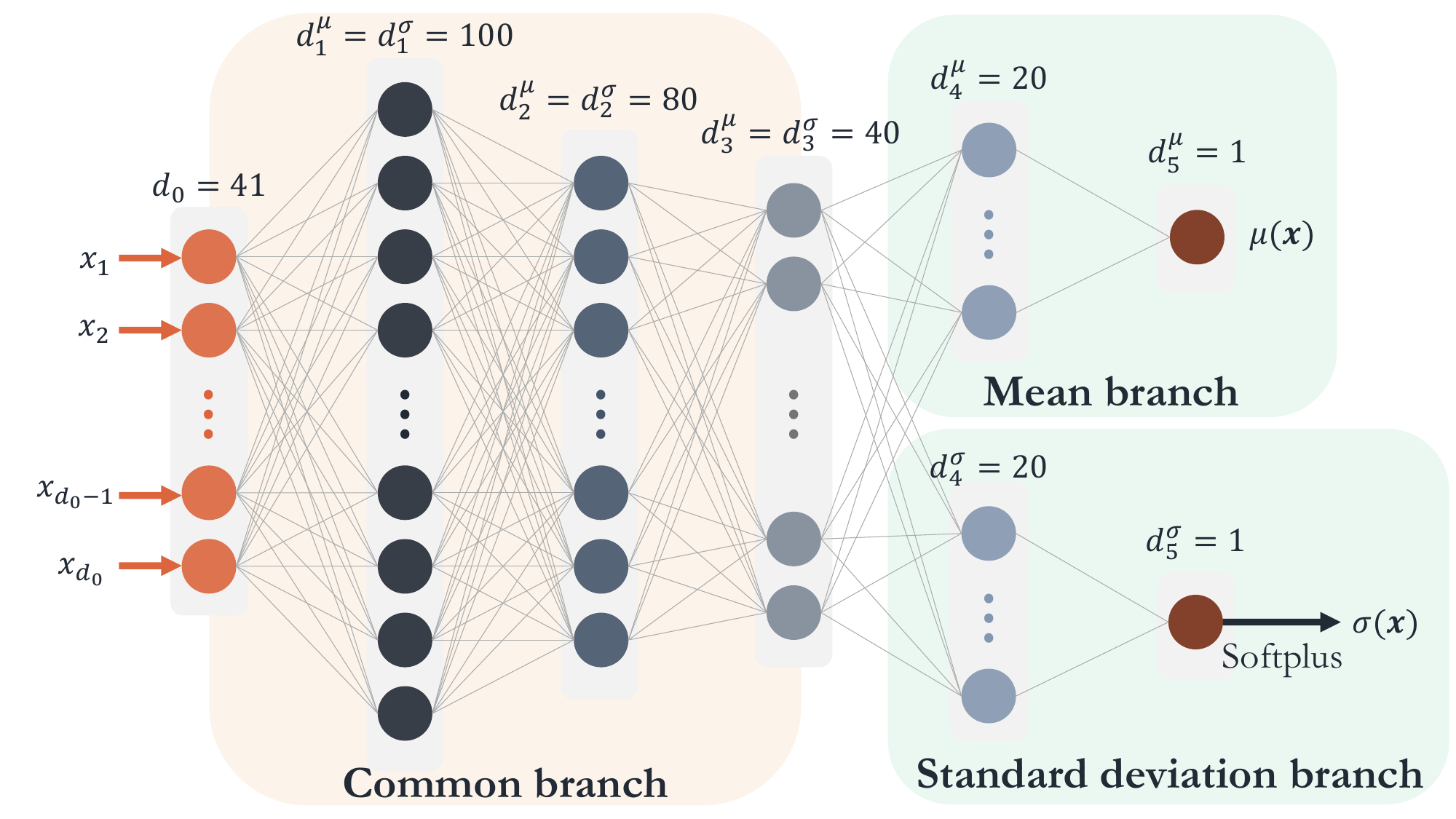}
    \caption{Architecture A1 used in the application of Section \ref{experiments}  for both PMLP and LPMPL with shared initial layers and independent branches for mean and standard deviation.}
    \label{fig:a1}
 \end{figure}

\begin{figure}[htbp]
   \centering
   \includegraphics[width=1\columnwidth]{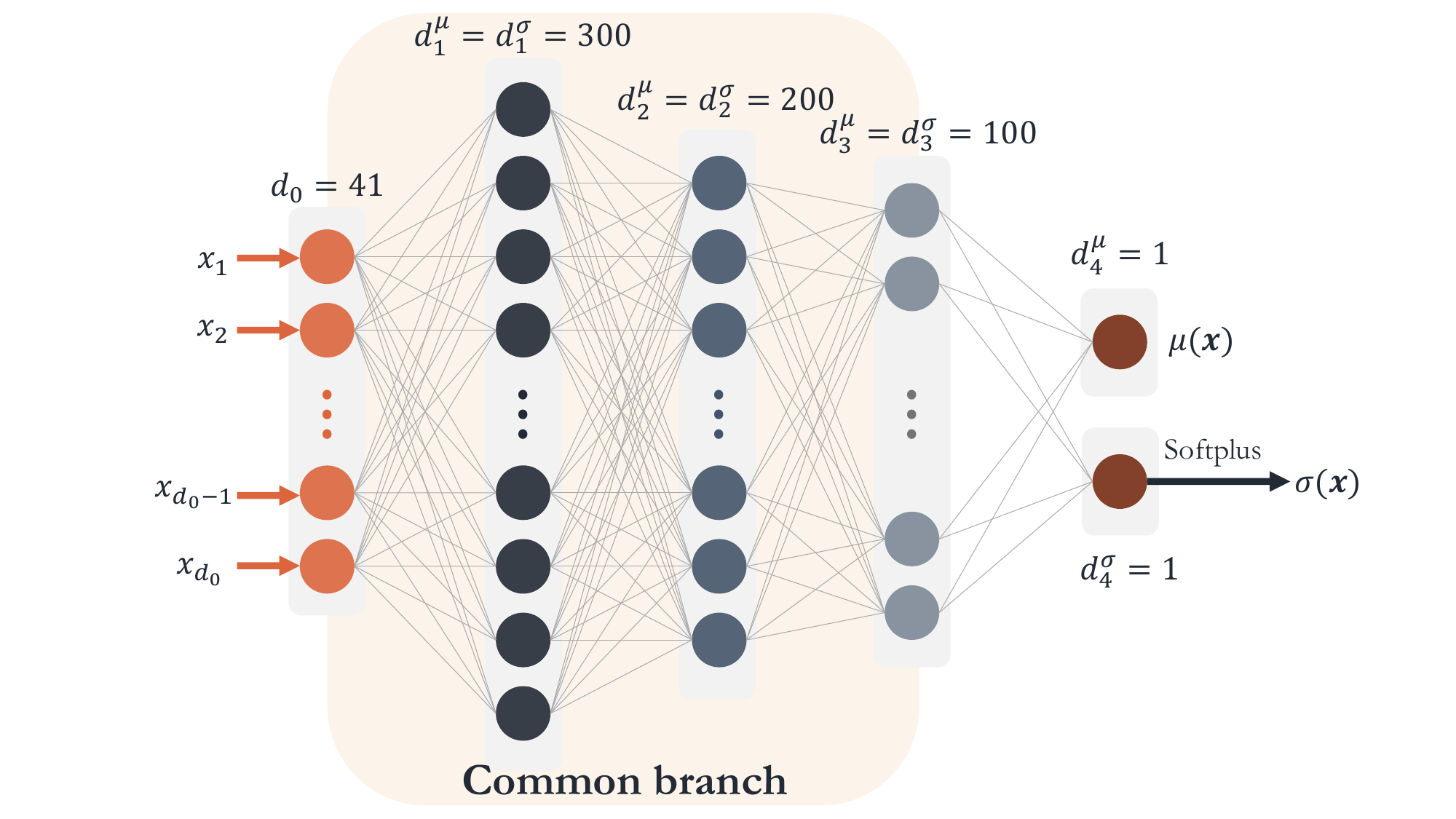}
   \caption{Architecture A2 used in the application of Section \ref{experiments}  for both PMLP and LPMPL with shared initial layers.}
   \label{fig:a2}
\end{figure}

In the application presented in Section \ref{experiments}, we compare two network architectures, A1 and A2. Architecture A1 is defined by $k_\mu=k_\sigma = 4$, $k^*= 3$, $d_0=41$, $d^\mu_1=d^\sigma_1 = 100$, $d^\mu_2=d^\sigma_2 = 80$, $d^\mu_3=d^\sigma_3 = 40$, $d^\mu_4=d^\sigma_4 = 20$, and architecture A2 by $k_\mu=k_\sigma = 3$, $k^*= 3$, $d_0=41$, $d_1^\sigma=d_1^\mu=300$, $d_2^\sigma=d_2^\mu=200$, $d_3^\sigma=d_3^\mu=100$. A visual representation of these architectures is provided in Figs. \ref{fig:a1} and \ref{fig:a2}, respectively. We
incrementally built these architectures by inspecting loss functions on training and testing data.

 The training of the PMLP is achieved by solving the following empirical risk minimisation problem.  Given $n$ data points $(\mathbf{x}_i,y_i) \in \Re^{d_0} \times \Re$, $i=1,2,\ldots,n$, and number and widths of the hidden layers, find ${\hat y}_i$ that is represented with an PMLP and minimises
 $
 \min_{\mathbf{W,b}} \mathcal{L}_1
 $
 where $\mathbf{W}=\{ \mathbf{W}_i \}_{i=1}^{k_{\text{total}}}$, $\mathbf{b}=\{ \mathbf{b}_i \}_{i=1}^{k_{\text{total}}}$ denoting all parameters, and $\mathcal{L}_1$ is a loss function.  By assuming independence between ${\hat y}_i$, the product of normal densities 
 $\prod_{i=1}^{n} \mathcal{N}(\mu(\mathbf{x}_i), \sigma^2(\mathbf{x}_i))$
 can be viewed as a multivariate predictive density so a plausible loss function for our PMLP is the minus logarithmic score, see for example  \cite{gneiting2007strictly},  defined as 
\begin{equation} 
 \label{eq:negloglike}
 \mathcal{L}_1 = -\sum_{i=1}^{n} \log \left( \mathcal{N}(y_i \mid \mu(\mathbf{x}_i), \sigma^2(\mathbf{x}_i)) \right) 
\end{equation}
where $\mathcal{N}(y | \mu,\sigma)$  denotes the p.d.f. of a normal density with mean $\mu$ and variance $\sigma^2$ evaluated at $y$.

\subsection{Transfer learning via fine-tuning} \label{prob-mlp-ft}
The realistic application of condition monitoring of a wind turbine should necessarily take into account the fact that the wind turbine is part of a wind farm. Thus, instead of training a model for each wind turbine separately, we could use the PMLP model of subsection \ref{prob-mlp} to train all wind turbines simultaneously. No particular care is needed other than treating the data so that the corresponding feature values of each wind turbine correspond to the power output of this wind turbine: our proposed PMLP treats the power  output conditional on the corresponding features of each $10$-minute period independently of the  power outputs of the other $10$-minute periods.  
Therefore, we can apply model (\ref{model}) to all wind turbines of the wind farm by just denoting $\mathbf{x}$ as the features from all wind turbines and ${\hat y}$ as a vector power output for a particular $10$-minute period. The loss function (\ref{eq:negloglike}) is just the sum of log-normal densities of each wind turbine at each $10$-minute interval. The mathematical formulation of this  model is as follows.  

Let  $\mathbf{x}_{ij}$ be the input feature vector of the $j$-th wind turbine  at the $i$-th $10$-minute interval with corresponding  output power $y_{ij}$. Then, the  model is defined as {\small
\begin{align}
 \mu(\mathbf{x}_{ij}) & = g^\mu_{k_\mu+1} \circ ReLU^\mu_{k_\mu} \circ g_{k_\mu}^\mu  \cdots  g^\mu_2 \circ ReLU^\mu_1 \circ g^\mu_1(\mathbf{x}_{ij}) \nonumber \\
 \sigma(\mathbf{x}_{ij}) & = S \circ g^\sigma_{k_\sigma+1} \circ ReLU^\sigma_{k_\sigma} 
 \circ g^\sigma_{k_\sigma}  
 \cdots   g^\sigma_2 \circ ReLU_1^\sigma \circ g^\sigma_1(\mathbf{x}_{ij}) \nonumber \\
 {\hat y}_{ij} & \sim \mathcal{N}(\mu(\mathbf{x}_{ij}), \sigma^2(\mathbf{x}_{ij}) ). \label{model2}
\end{align}}
If we have data for $n_j$ time intervals for wind turbine $j$ for $j=1,\ldots,J$, the  loss function of model (\ref{model2}) is
\begin{equation} 
 \label{eq:negloglike2}
 \mathcal{L}_2 = - \sum_{j=1}^{J}  \sum_{i=1}^{n_j}  \log \left( \mathcal{N}(y_{ij} \mid \mu(\mathbf{x}_{ij}), \sigma^2(\mathbf{x}_{ij})) \right).
\end{equation}
Notice that the possibly different number of available data $n_j$  from each wind turbine indicates that this model utilizes all available data from all wind turbines.  This is of huge practical importance in realistic applications since SCADA data typically contain many missing data.

However, by training all data of wind farms in the same model, the amount of available data may become prohibitively vast to allow frequent re-training.  The usual treatment of such huge datasets is to adopt some transfer learning techniques so that knowledge gained from pre-training the model based on all wind turbines data can be used to boost performance in predicting the power of a single wind turbine.   We call such a PMLP model that uses pre-training based on loss function $\mathcal{L}_2$  and then predicts the power of only one wind turbine using fine-tuning and the loss function $\mathcal{L}_1$ 
a Large PMLP (LPMLP) model.
When new data arrive, re-training of the PMPL model based on $\mathcal{L}_2$ is not anymore necessary. Moreover, additional important practical advantages arise.  First, consider the very realistic scenario in which 
a wind turbine has fewer data points because, for example, it has been out-of-order for a long period of time or has been recently installed. Normal behaviour modelling of this wind turbine might be very hard or even impossible to achieve.  Our LPMLP model is capable of producing a  predictive density for such low-information wind turbines by using the data from all wind turbines in the wind farm.
Second, transfer learning can be used to improve the predictive power of  model (\ref{model}) as we empirically show in Section \ref{experiments}.

Our proposed transfer learning is achieved via the following fine-tuning.  We first train the LPMLP model (\ref{model2}) with the loss function (\ref{eq:negloglike2}).
Then, we use the parameters of the pre-trained model as initial values, and we  predict the mean and standard deviation of the output power of the wind turbine we are interested in.  Thus, the training uses the loss function (\ref{eq:negloglike}) that refers to one only wind turbine and is achieved very fast. {We apply this training method to both the A1 and A2 architectures to investigate whether additional parameters, combined with a pre-training phase, provide greater benefits to the model compared to pre-training alone. Specifically, architectures A1 and A2 differ in their branching and number of parameters, which are $16,508$ and $91,002$, respectively.}

For the advantages of transfer learning in similar problems see, for example, \cite{ma2024transfer}. Notable theoretical and empirical motivations for introducing transfer learning in the given probabilistic regression context include (i) regularisation:  pre-training the LPMLP allows learning general features and patterns from a larger amount of data which can be especially beneficial in the case where the operating conditions of the wind farm are extremely diverse.  Thus, it is easier for the model to deal with conditions that have not been observed in the particular wind turbine under inspection; (ii) data sparsity: pre-training on the whole data of a wind farm dataset can help alleviate data sparsity by providing additional data for learning more general but not yet observed representations. For example, see  \cite{liao-2023} for a discussion of limited data availability for newly built wind farms; and (iii) improved convergence: pre-training the LMPLP  facilitates faster converge during fine-tuning on a single wind turbine. 
\subsection{Condition Monitoring}\label{monitoring}
The reason that we propose the PMLP and LPMLP models which have a probabilistic flavour by predicting both the mean and the variance of the output power is to construct a reasonable, probability-based, data-driven condition monitoring system.  This system 
has the usual ingredients that consist of first training the model during a healthy wind turbine operation and then observing both SCADA and output power data.  Anomalous operational behaviour is reported when the observed output power departs, in a probabilistic sense, from the expected predicted normal density under the trained model. Our condition monitoring system is based on testing model adequacy by testing the hypothesis that the observed output power observations have been generated by our trained model.  This requires, of course, some evidence that our model has sufficiently good coverage probability so that in an out-of-sample healthy wind turbine environment the observed power output indeed follow the normal density indicated by the trained model.  As it will be evident in our application study, our proposed PMLP and LPMLP models  do indeed have good coverage probabilities in an out-of-sample large empirical exercise.

Assume that the model training phase has ended and we are now in the live condition monitoring phase. To emphasize that the data is observed sequentially in time,  we will be now using a subscript $t$ rather than $i$ for the time intervals. Assume that we observe $T$ consecutive power outputs $y_t,\;t=1,2,\ldots,T$, and we have predictions from our model $N(\mu(\mathbf{x}_t),\sigma^2(\mathbf{x}_t))$.  Under the hypothesis of model adequacy, or no-fault, we define  $v_t = (y_t - \mu(\mathbf{x}_t))/\sigma(\mathbf{x}_t$) and test whether the sample $v_t,\;t=1,2,\ldots,T$ comes from a $N(0,1)$ distribution.  {For example, we could perform a test over seventy-two hours with 10-minute intervals, where $T=432$.} 

We propose a two-sided tabular CUSUM control chart \citep{montgomery2009statistical} defined as $$S_{\mathrm{H}}(t)=\max \left\{0, v_{t}-k+S_{\mathrm{H}}(t-1)\right\}$$ and $$S_{\mathrm{L}}(t)=\max \left\{0,-k-v_{t}+S_{\mathrm{L}}(t-1)\right\},$$ where $S_{\mathrm{H}}(0)=S_{\mathrm{L}}(0)=0$.  The reference (or allowance or slack) value of $k$ is traditionally specified with past experience or from simulations based on supervised settings. A common choice is about halfway between the target mean zero and the out-of-control value that we are interested in detecting quickly.  Therefore, if we are interested in detecting a deviance of one standard deviation a plausible choice is $k=1/2$; see \cite{montgomery2009statistical}.  Recent advances propose an adaptive adjustment of its value \citep{wu2009enhanced}. Note that  $S_{\mathrm{H}}(t)$ and $S_{\mathrm{L}}(t)$ are viewed as cumulative deviations from zero that are greater than $k$ that are not allowed to take negative values.  If either of them exceeds a decision interval $I$, the process is considered to be out-of-control.  A reasonable choice of $I$ is five times the process standard deviation \citep{montgomery2009statistical} which is one in our case, {suggesting $I=5$.  In practice, to define the decision interval $I$ according to the needs of the wind farm, $I$ can be estimated from empirical data to achieve the desired combination of recall and precision. In Section~\ref{sensitivity analysis}, we present an illustration in which we evaluate the model's performance in terms of precision and recall. These metrics are defined as 
as $\textit{precision}=\textit{TP}/(\textit{TP}+\textit{FP})$ and $\textit{recall}=\textit{TP}/(\textit{TP}+\textit{FN})$
where $\textit{TP}$, $\textit{FP}$, and $\textit{FN}$ denote the number of true positives, false positives, and false negatives, respectively. }  For $t=1,\ldots,T$ the test statistics are defined as $A_{t}^{\mathrm{H}}=$ $\max \left\{S_{\mathrm{H}}(i), i=1, \ldots, t\right\}$ and $A_{t}^{\mathrm{L}}=\max \left\{S_{\mathrm{L}}(i), i=1, \ldots, T\right\}$. The null hypothesis of normal operation in the wind turbine is rejected (with a one-sided test) if $A_{n}=\max \left(A_{t}: 1 \leqslant t \leqslant n\right)$ exceeds its upper critical value, where $A_{t}=\max \left(A_{t}^{\mathrm{H}}, A_{t}^{\mathrm{L}}\right)$. 

\section{Experiments} \label{experiments}
\subsection{Data Acquisition and Filtering} \label{data}
The application utilises $10$-minute SCADA and events data from the six Senvion MM92 wind turbines at Kelmarsh wind farm in the UK \citep{data}. The dataset spans from $3^{\text{rd}}$ January $2016$ to $1^{\text{st}}$ July $2021$ and comprises over $1.7$ million data points containing $110$ variables including date-time, wind speed, bearing temperature, and power output. The data includes $10$-minute averages, standard deviations,  minimum and maximum values of all measured variables. The recording of wind speed summary statistics started at $25^{\text{th}}$ September $2017$ so we considered only data entries from this date onwards.

\begin{figure}[ht]
   \centering
   \includegraphics[width=1\columnwidth]{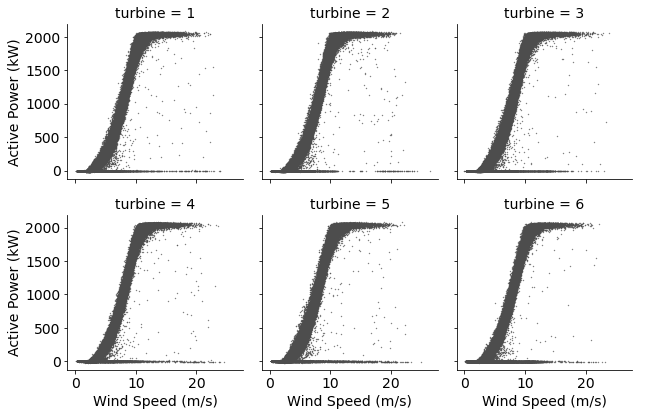}
   \caption{Real data wind power curves for all six wind turbines at Kelmarsh wind farm before removing standbys and warnings existing in the operational status and event file.}
   \label{fig:outliers_before}
\end{figure}

\begin{figure}[ht]
   \centering
   \includegraphics[width=1\columnwidth]{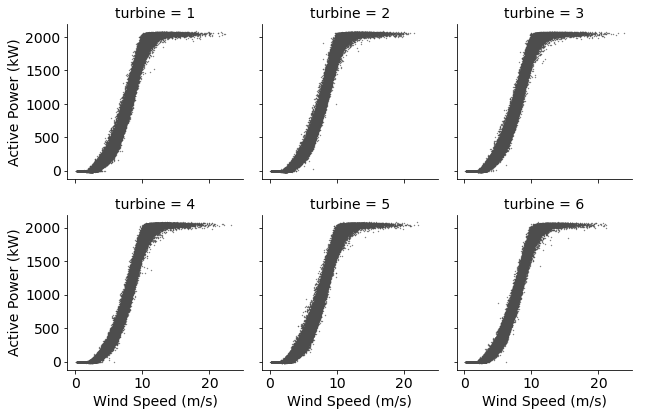}
   \caption{Real data wind power curves for all six wind turbines at Kelmarsh wind farm after removing standbys and warnings using the operational status and events file provided by the data provider \cite{data}.}
   \label{fig:outliers}
\end{figure}

An operational status and events file was utilised for data filtering to ensure consistent modelling of normal behaviour. These files provide valuable insights into the operational conditions of the wind turbines, covering a range of scenarios from technical failures to operational or environmental standbys and warnings. To enable accurate behaviour modelling, it was essential to remove out-of-control condition data records and base our model training solely on filtered data.  {For filtering, periods of standby, warnings, and operational stops were excluded. Furthermore, data from the week preceding each forced outage was removed to reduce the likelihood of including out-of-control events in the training set.}
The data elimination process is illustrated in Fig. \ref{fig:outliers_before} and \ref{fig:outliers} where plots of output powers against wind speeds are depicted for each wind turbine before and after the filtering.

All our analyses used as input features $41$ operational and environmental variables are listed in Table \ref{tab:feature_statistics}. The final filtered dataset across the six wind turbines comprises {$846,968$ data points, with $163,562$ of them belonging to one single wind turbine} used in the application Section \ref{app}. 

\begin{table}[h!]
\centering\caption{SCADA operational and environmental variables used as input features in the proposed models. Avg: Average; Stdev: Standard deviation; Min: Minimum; Max: Maximum; temp: temperature; {dir: direction}.}
\begin{tabular}{ll}
\toprule \textit{Feature description} & \textit{Feature description }\\\midrule\midrule
Avg. wind speed & Avg. front bearing temp. \\
Stdev. wind speed & Stdev. front bearing temp. \\
Min. wind speed & Min. front bearing temp. \\
Max. wind speed  & Max. front bearing temp. \\
Avg. rear bearing temp. & Avg. rotor bearing temp. \\
Stdev. rear bearing temp. & Avg. stator1 temp. \\
Min. rear bearing temp. & Avg. nacelle ambient temp.\\
Max. rear bearing temp. & Avg. nacelle temp. \\
Avg. transformer temp. & Avg. gear oil temp. \\
Avg. gear oil inlet temp. & Avg. drive train acceleration \\
Avg. top box temp. & Avg. hub temp.\\
Avg. conv. ambient temp.& Avg. transformer cell temp.\\
Avg. motor axis1 temp. & Avg. motor axis2 temp. \\
Avg. CPU temp. & Avg. blade angle pitch A \\
Avg. blade angle pitch B & Avg. blade angle pitch C  \\
Avg. gear oil inlet press & Avg. gear oil pump press  \\
Tower acceleration x &Tower acceleration y\\
Sine avg. wind speed dir. & Sine max. wind speed dir.\\
Cosine avg. wind speed dir. & Cosine max. wind speed dir.\\
Sine min. wind speed dir. & Stdev. wind speed dir.\\
Cosine min. wind speed dir.\\

\bottomrule
\end{tabular}
\label{tab:feature_statistics}
\end{table}

\subsection{Application}\label{app}

For all our experiments, we used a {chronologically} selected $80-20$ data ratio for the train and test data, respectively. When a validation set was required, the train-validation data ratio was $72-8$.  {To select the validation set, the training set was initially shuffled to ensure that data from different seasons were included in the validation set.} The PMLP and LPMLP models were trained using the Adam optimiser with a learning rate of $0.001$, a batch size of $32$,  and $100$ and $500$ epochs respectively. This was done for both A1 and A2 architectures. Additionally, the LPMLP models were fine-tuned for 50 epochs using learning rates of $10^{-3}$ and $10^{-4}$ for A1 and A2 respectively. The Softplus activation function (\ref{softplus}) used a $\delta=0.001$. Early stopping on a validation set was employed during all training stages. All results refer to the out-of-sample prediction of the output power of only one randomly chosen wind turbine of the wind farm.

To evaluate the proposed probabilistic PMLP and LPMLP modelling frameworks, we compared their performance to other probabilistic methods. In particular, we used a Gaussian process regression and a Bayesian neural network with stochastic output.  The  Gaussian process inference was based on variational inference with inducing points \citep{Snelson} using a radial basis function (RBF) kernel and $100$ pseudo-inputs. This approximate inference procedure was used because the large sample size required low computational complexity methodologies, a problem quite useful in Gaussian process literature; see, for example, \cite{titsias2009variational}.  { In particular, the time complexity goes from $\mathcal{O}(n^3)$ of traditional Gaussian processes to $\mathcal{O}(n\cdot m^2)$ when using variational inference with inducing points, where m is the number of inducing points such that $m << n$.}
The Bayesian Neural Network was defined as a set of fully connected layers with mean field normal hierarchical priors 
with mean and standard deviation having  $\mathcal{N}(0, 0.1)$ and  $\mathcal{N}(S(0.001), 0.1)$ 
hyperpriors respectively; here, $S$ is the Softplus function defined in (\ref{softplus}) with $\delta = 0$.  It was trained using the Flipout method which re-models the stochastic variational inference implementation strategy as a weight perturbation \citep{50019} over $100$ epochs using early stopping with the loss given by  (\ref{eq:negloglike}).  {The time complexity of training Bayesian neural networks and PMLPs is upper bounded by $\mathcal{O}(n \cdot e \cdot k)$ where $n$ is the number of training data points, $e$ is the number of epochs trained and $k$ is the number of neurons in the networks. }

\begin{figure}[ht]
   \centering
   \includegraphics[width=1\columnwidth]{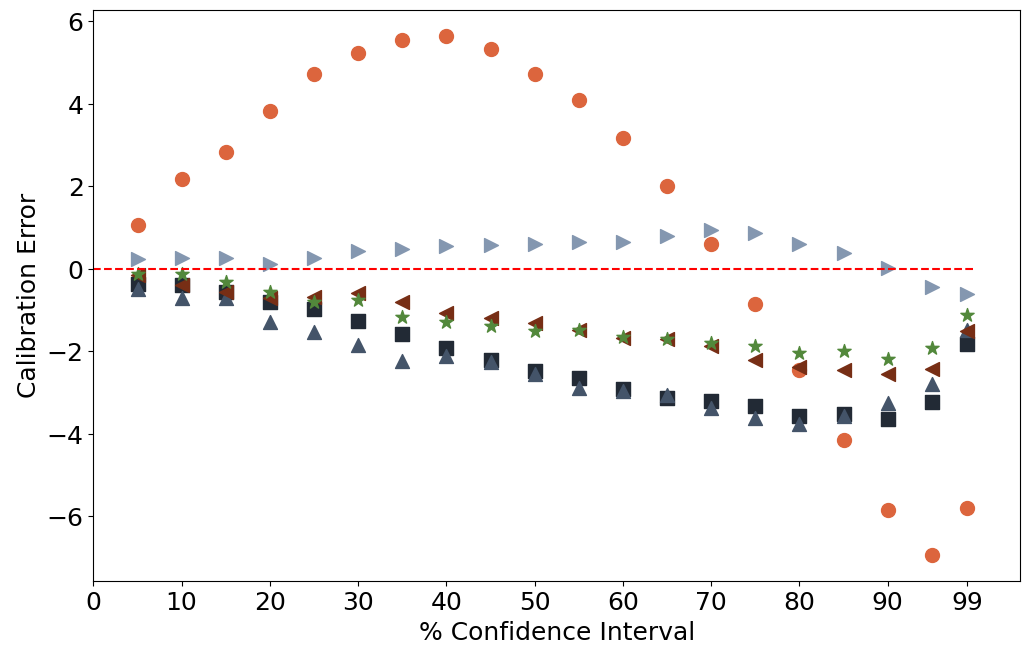}
   \caption{
   Calibration error of twenty binned confidence levels including the 95\% and 99\% confidence intervals. \textcolor{color1}{$\bullet$} Sparse Gaussian process (RBF kernel); \textcolor{color3}{$\blacktriangle$} PMLP with architecture A1;  \textcolor{color5}{$\blacktriangleleft$} PMLP with architecture A2;  \textcolor{color2}{$\blacksquare$} Bayesian neural network; \textcolor{color4} {$\blacktriangleright$} LPMLP with architecture A1; \textcolor{color6}{$\bigstar$} LPMLP with architecture A2.}
   \label{fig:summary}
\end{figure}

The results shown in Table \ref{table:results} are based on { $130,849$ and $32,713$ training and test $10$-minutes intervals respectively}.  Note that the models have been trained with SCADA data which have the size of nearly three years of wind turbines operation.    We base our comparison with the usual root mean square and mean absolute error defined as $\text{RMSE} = (n^{-1} \sum_{i=1}^{n} (y_i - \mu(\mathbf{x}_i))^2)^{1/2}$ and 
$\text{MAE} = n^{-1} \sum_{i=1}^{n} |y_i - \mu(\mathbf{x}_i)|$ respectively. {We also present the RMSE and MAE metrics normalised by the rated power (2050 kW) to provide a clearer assessment of algorithm performance.}  Although these metrics provide a good indication on how well our model predicts the output power, it is based only on the predicted means $\mu(\mathbf{x}_i)$.  An issue of more importance to our condition monitoring methodology is the out-of-sample coverage probabilities that are defined as the probabilities that a confidence interval region will include the true power output.  These are empirically estimated in our test data and  shown in Fig. \ref{fig:summary} where the calibration error, defined as the difference  between observed and theoretical coverage probabilities are plotted for all six models and for twenty different intervals.  { The maximum calibrated error, representing the maximum observed deviation, along with the 95\% and 99\% coverage probabilities, is reported in Table \ref{table:results}.}

\begin{table*}[!ht]
    \centering
    \caption{Out-of-sample performance metrics {related to the prediction of active power output (measured in kW). The units for each metric are indicated in parentheses next to the metric names. PMLP: probabilistic multi-layer perceptron; LPMLP: large probabilistic multi-layer perceptron; RMSE: root mean square error; MAE: mean absolute error; NRMSE: normalised root mean square error; NMAE: normalised mean absolute error; MCE: maximum calibration error; CP: coverage probability. Best performance is indicated in bold. NRMSE and NMAE are expressed as percentages of the rated power (2050 kW).}}
    \begin{tabular}{l|cc|cc|ccc}
        \toprule
         \textit{Methods} & \textit{RMSE (kW)} & \textit{MAE (kW)} & \textit{NRMSE (\%)} & \textit{NMAE (\%)} & \textit{MCE (\%)} & \textit{95\% CP (\%)} & \textit{99\% CP (\%)} \\ \midrule\midrule
        \textit{Sparse Gaussian process}   & $49.57$ & $33.87$ & $2.42$ & $1.65$ & $6.94$ & $88.03$ & $93.20$ \\ 
        \textit{Bayesian neural network}   & $40.28$ & $22.54$ & $1.96$ & $1.10$ & $3.63$ & $91.79$ & $97.18$ \\ 
        \textit{PMLP -- A1 architecture} & $28.97$ & $15.94$ & $1.41$ & $0.78$ & $3.74$ & $92.21$ & $97.52$ \\ 
        \textit{PMLP -- A2 architecture} & $28.84$ & $15.94$ & $1.41$ & $0.78$ & $2.54$ & $92.57$ & $97.48$ \\ 
        \textit{LPMLP -- A1 architecture} & $\mathbf{27.38}$ & $\mathbf{14.81}$ & $\mathbf{1.34}$ & $\mathbf{0.72}$ & $\mathbf{0.93}$ & $\mathbf{94.56}$ & $\mathbf{98.40}$ \\ 
        \textit{LPMLP -- A2 architecture} & $27.52$ & $14.91$ & $1.34$ & $0.73$ & $2.17$ & $93.09$ & $97.89$ \\ 
        \bottomrule
    \end{tabular}
    \label{table:results}
\end{table*}

{Note that the architectures of A1 and A2 in the PMLP and LPMLP models differ, as described in Section \ref{prob-mlp} and illustrated in Figs. \ref{fig:a1} and \ref{fig:a2}. By employing these two architectures, which were designed based on an inspection of the loss functions during training and testing, we investigated the influence of the number of layers and neurons, branching, and activation functions.} For our empirical exercise, it is of particular interest to investigate whether the scaling of the architecture or the transfer learning is the actual cause of the  model performance improvement.  { To illustrate this, Table \ref{table:results} and Fig. \ref{fig:summary} present the results of both the PMLP and LPMLP models with architectures A1 and A2.} 

By controlling the network architecture, there is evidence that pre-training improves both RMSE and MAE metrics. Additionally, our proposed configuration outperforms the sparse Gaussian process across all metrics while significantly reducing training times. For instance, using an Intel(R) Core(TM) i7-10510U CPU, the sparse Gaussian process required $168$ minutes of training, compared to only $6$ minutes for the Bayesian neural network. The PMLP with architecture A1 required $35$ minutes, while the PMLP with architecture A2 completed training in $25$ minutes. Similarly, the LPMLP with architecture A1 underwent pre-training for $79$ minutes and fine-tuning for $8$ minutes, whereas the LPMLP with architecture A2 was pre-trained for $103$ minutes and fine-tuned for $2$ minutes.

\begin{figure}[h!]
   \centering
   \includegraphics[width=1\columnwidth]{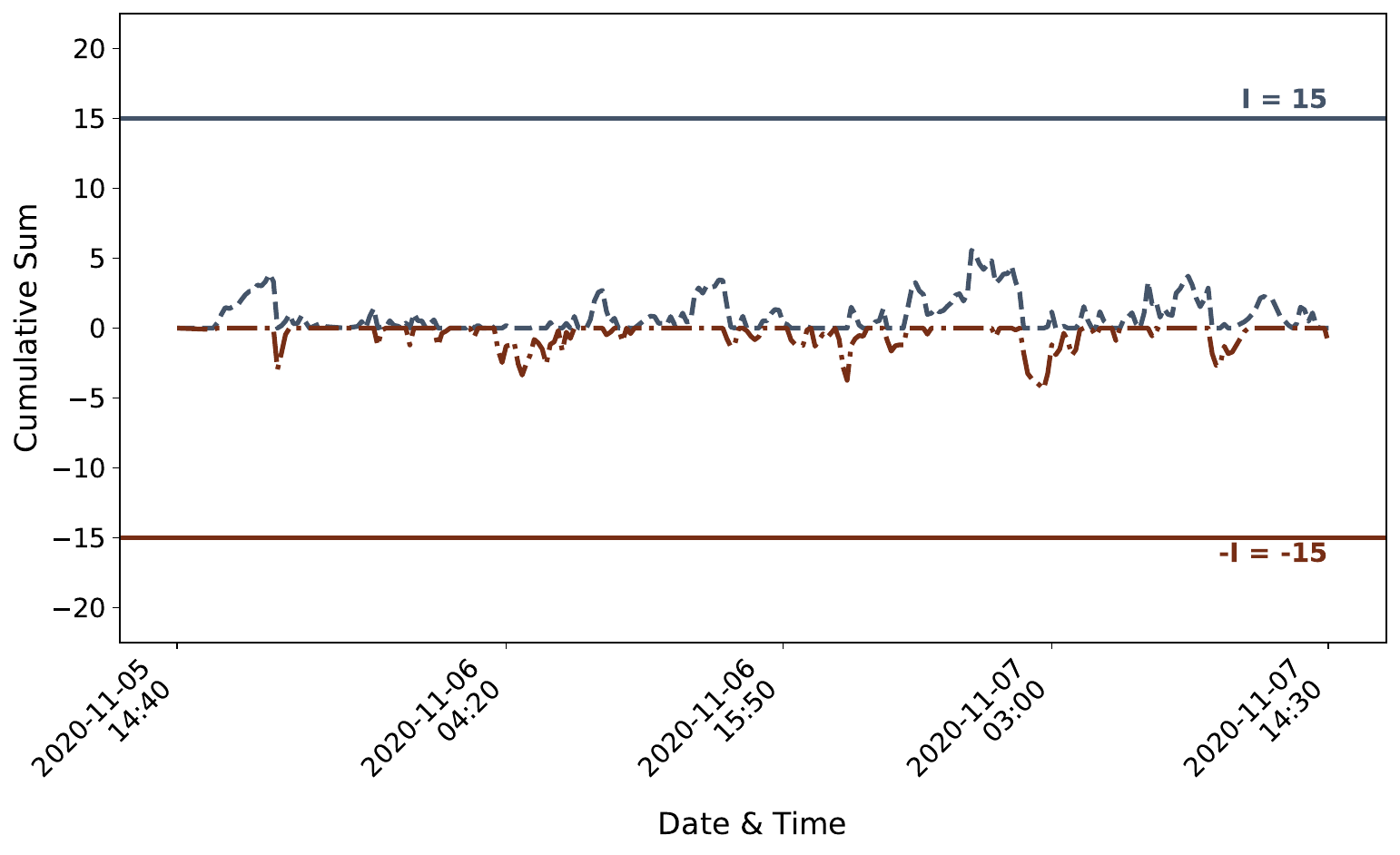}
   \caption{CUSUM control chart for seventy-two hours during normal wind turbine behaviour. In the chart (\protect\myline{color3}), (\protect\myline{color5}), (\protect\mydashline{color3}), and (\protect\mydashdotline{color5}) represent $I$, $-I$, $S_{H}$ and $-S_{L}$.
    }
   \label{fig:cusum_healthy}
\end{figure}

\begin{figure}[h!]
   \centering
   \includegraphics[width=1\columnwidth]{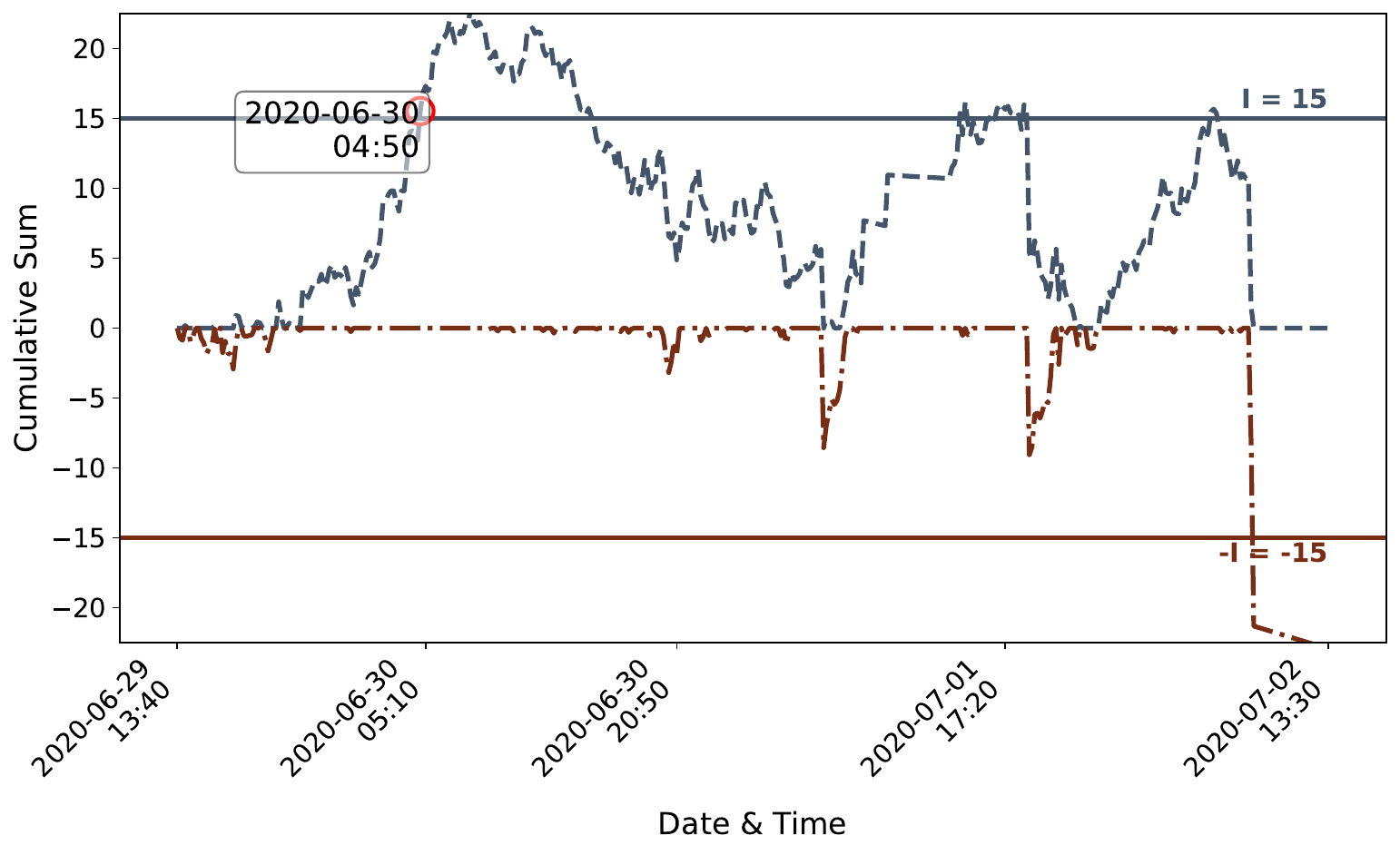}
   \caption{CUSUM control chart for seventy-two hours prior to an operational fault started on the $\text{2}^{\text{nd}}$ of July $2020$ at $13$:$35$PM. In the chart (\protect\myline{color3}), (\protect\myline{color5}), (\protect\mydashline{color3}), and (\protect\mydashdotline{color5}) represent $I$, $-I$, $S_{H}$ and $-S_{L}$.   The process starts to be out-of-control for the first time on $\text{30}^{\text{th}}$ of June $2020$ at $04$:$50$AM.
    }
   \label{fig:cusum_unhealthy}
\end{figure}

We illustrate our proposed unsupervised method for condition monitoring using two different examples, one in which the examined wind turbine is operating normally and one in which the process is observed to be out-of-control;  the results of these examples are presented in the CUSUM control charts of Fig. \ref{fig:cusum_healthy} and \ref{fig:cusum_unhealthy} respectively.  In both cases, observations obtained over seventy-two hours, consisting of $T=432$ $10$-minute intervals, are monitored using CUSUM control charts.  Fig. \ref{fig:cusum_healthy} presents a typical situation of a good operation of a wind turbine.    

In contrast, Fig. \ref{fig:cusum_unhealthy}  shows that the CUSUM control chart indicates the process is out-of-control { for the first time at $04$:$50$AM on the $\text{30}^{\text{th}}$ of June $2020$.  By inspecting the operational status and event file of this wind turbine, we identified a triggered alarm on the $\text{2}^{\text{nd}}$ of July $2020$ at $13$:$35$PM, related to a forced outage caused by a converter error. }

\subsection{Choice of decision interval}
\label{sensitivity analysis}

A practical way to select the decision interval $I$ for the CUSUM control charts is to inspect historical data and choose an $I$ value that achieves the desired balance between false negatives and false positives. Unfortunately, such analysis is not possible for our experimental data because a detailed list of operational anomalies is unavailable: out of approximately $163$K recordings, only twenty-two faults have been reported. However, we present here a pathway for selecting $I$. This analysis is based on the LPMLP architecture A1, using precision and recall metrics to evaluate the effectiveness of fault detection.

\begin{figure}[htbp]
   \centering
   \includegraphics[width=1\columnwidth]{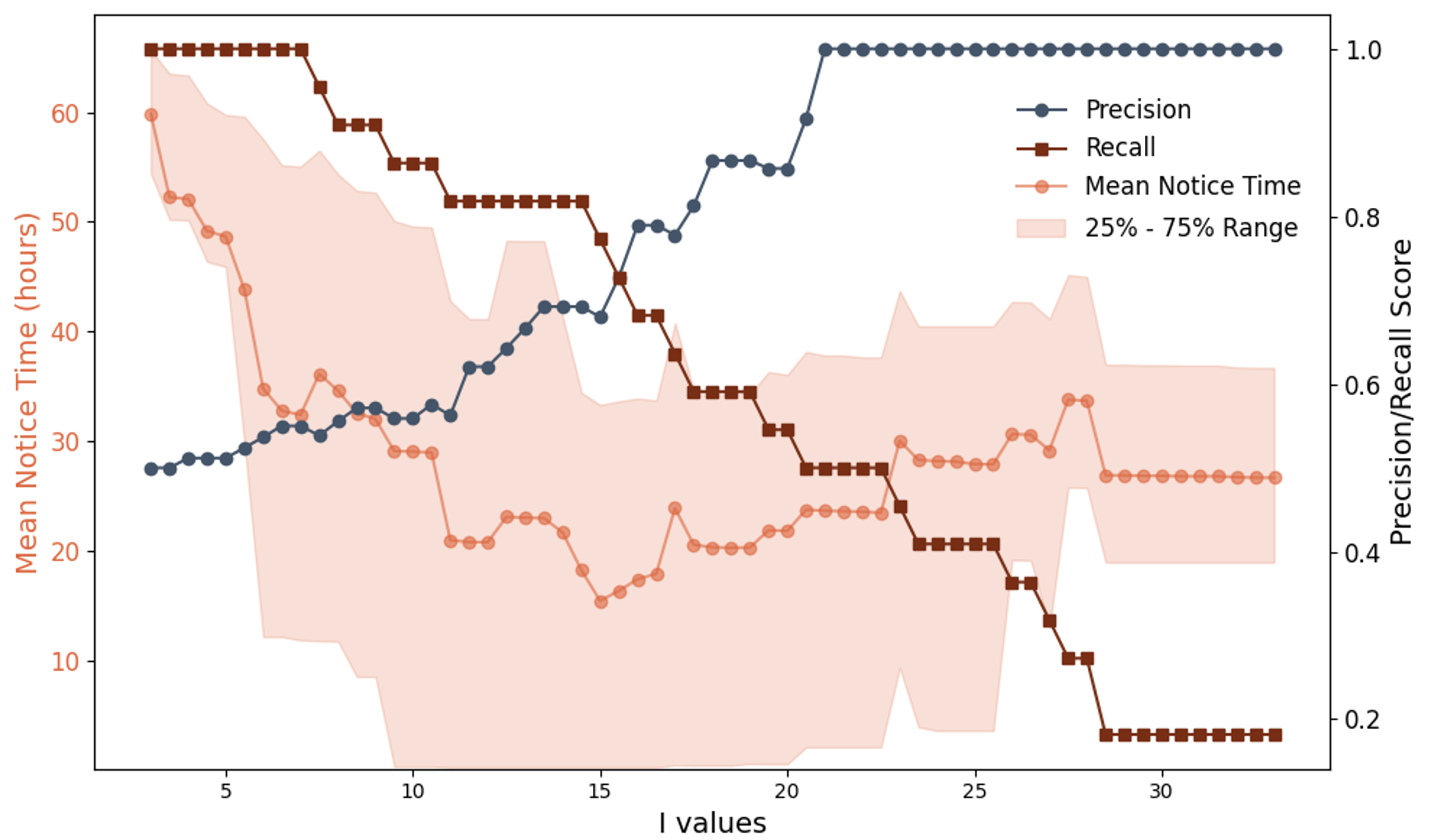}
   \caption{ Precision and recall scores for the LPMLP architecture A1 performance with varying parameter $I$ values. The plot illustrates the performance metrics alongside the mean notice time in hours for fault cases.  The shaded area represents the first and third quartiles of notice times for different values of $I$.}
   \label{fig:recall-precision}
\end{figure} 

We evaluated the performance of the LPMLP model across different values of $I$. To obtain balanced data for our classification exercise, we used twenty-two randomly selected seventy-two-hour periods of normal operation and twenty-two seventy-two-hour periods during which forced outages were recorded in the operational status and event files. The forced outages were caused by various factors, the most common being tower oscillation faults, which could potentially lead to structural damage and errors in the generator-converter system. Less commonly observed issues included yaw and/or fan overload faults caused by overheating or excessive electrical load, as well as faults due to malfunctions in pitch control systems, which could result in potential blade damage and mechanical stress.

The results for the precision and recall metrics for different values of $I$ are presented in Fig. \ref{fig:recall-precision}. A visual inspection of this plot indicates the optimal choice of the decision interval, depending on the desired sensitivity of the monitoring system to precision and recall values. For example, based on Fig. \ref{fig:recall-precision}, a good compromise is the value $I=15$, where the precision is $0.68$ and the recall is $0.77$. This successfully identifies, in the out-of-sample test data, seventeen out of twenty-two instances of faults. Notably, the average lead notice time for anomaly detection by our monitoring system is $15.49$ hours, with a standard deviation of $21.76$ hours.

\section{Discussion}
\label{discussion} 

When the condition monitoring system identifies malfunction, such as the one in Fig. \ref{fig:cusum_unhealthy}, the engineers in the field would like to identify which component of the wind turbine is causing the problem. To achieve this, we could extend our method to monitor different variables of the SCADA inputs, such as temperatures or pressures across various parts of the wind turbine. An obvious way to achieve this is to replicate our methodology by replacing the output power with another variable.  A complete condition monitoring system would thus be a collection of CUSUM control charts with different outputs modelled with different LPMLP models.

A further methodological direction would be to replace the fine-tuning of LPMLP which is based on parameter initialisation; see, for example, \cite{shwartz2022pre}.  We could use the parameter estimates of a pre-trained model to construct informative prior distributions for transfer learning. Then, highly informative posteriors will be available for the model that predicts the power of one single wind turbine.  This is, in essence, a Bayesian neural network with informative priors that replace the vague priors we used in our application.

\section{Conclusion}\label{conclusion}
We presented a condition monitoring system that can be immediately applied to a wind farm.  We took particular care to accommodate issues that are routinely met in pragmatic wind farm operations. In particular, all SCADA data from all wind turbines collected in  many years can be  used to train our model.  Our monitoring system has the ability to incorporate wind turbines that have many missing data.  It is based on a model that produces heterogeneous predictive densities that are well suited to wind turbine data; its probabilistic nature provides a scientifically sound monitoring system that produces automatic monitoring of a wind farm in the form of a control chart. 

For the data we used, our probabilistic model outperforms other probabilistic methods in terms of RMSE and MAE and has good coverage probabilities in large out-of-sample empirical exercises.  Although it is an unsupervised monitoring system, we were able to illustrate its immediate applicability by inspecting the fault events file after an anomaly is detected.

Future research involves investigating ways to incorporate a further step that will identify which operational characteristic of the wind turbine caused the anomaly detected by the control chart.

\section*{Acknowledgments}
We would like to thank Dr Evangelos Morfiadakis, CEO at International Wind Engineering, for introducing us to the problem and providing valuable insight; and Dr Charlie Plumley, Senior Performance Engineer at Glennmont Partners, for  helpful discussions. 
\bibliography{refs}

\end{document}